\documentclass[sigconf, authorversion, nonacm]{acmart}

\usepackage{times}
\usepackage{latexsym}
\usepackage{subcaption}
\usepackage{multirow}
\usepackage{multicol}
\usepackage{graphicx}
\usepackage{siunitx}
\usepackage{makecell}
\usepackage{color}
\usepackage{tabularx}
\usepackage{booktabs}

\AtBeginDocument{%
  \providecommand\BibTeX{{%
    \normalfont B\kern-0.5em{\scshape i\kern-0.25em b}\kern-0.8em\TeX}}}

\begin{document}

\title{Unveiling Global Narratives: A Multilingual Twitter Dataset of News Media on the Russo-Ukrainian Conflict}

\author{Sherzod Hakimov}
\email{sherzod.hakimov@uni-potsdam.de}
\affiliation{
  \institution{Computational Linguistics, Department of Linguistics\\University of Potsdam}
    \country{Germany}
}

\author{Gullal S. Cheema}
\email{gullal.cheema@tib.eu}
\affiliation{
  \institution{L3S Research Center\\Leibniz University, Hannover}
    \country{Germany}
  }

\renewcommand{\shortauthors}{Hakimov and Cheema}

\begin{abstract}

The ongoing Russo-Ukrainian conflict has been a subject of intense media coverage worldwide. Understanding the global narrative surrounding this topic is crucial for researchers that aim to gain insights into its multifaceted dimensions. In this paper, we present a novel multimedia dataset that focuses on this topic by collecting and processing tweets posted by news or media companies on social media across the globe. We collected tweets from February 2022 to May 2023 to acquire approximately 1.5 million tweets in 60 different languages along with their images. Each entry in the dataset is accompanied by processed tags, allowing for the identification of entities, stances, textual or visual concepts, and sentiment. The availability of this multimedia dataset serves as a valuable resource for researchers aiming to investigate the global narrative surrounding the ongoing conflict from various aspects such as who are the prominent entities involved, what stances are taken, where do these stances originate from, how are the different textual and visual concepts related to the event portrayed.

\end{abstract}



\maketitle

\section{Introduction}
The Russo-Ukrainian conflict, which began in February 2022, has been a focal point of global attention. The conflict has been extensively covered by news media channels worldwide, each presenting a unique perspective shaped by their geographical location, political stance, and cultural context. This extensive coverage has resulted in a wealth of data that, if properly harnessed, can provide valuable insights into the global perception and narrative of the conflict. However, to date, no comprehensive study has been conducted to analyze the coverage of the Russo-Ukrainian conflict by news media channels on social media, specifically focusing on multimedia data that does not only include text but also images. This gap in the literature motivates the need for a dataset that encompasses tweets and their images from news media channels across the globe, focusing on the topic of the Russo-Ukrainian war. Specifically, the analysis of discourse on social media from various perspectives such as within a certain country and news media companies. Currently, existing datasets about the conflict focused either more on the collection of language-specific subsets~\cite{park2022voynaslov, vahdat2023russia,toraman2022good},  applying only sentiment analysis methods~\cite{caprolu2022characterizing, shevtsov2022twitter, xu2023sentiment, dvzubur2022semantic}, or use multimedia data (tweet text and image) for down-stream tasks such as hate speech detection~\cite{thapa-etal-2022-multi,DBLP:conf/cvpr/BhandariSTNN23}. However, none of these previous studies focused specifically on the coverage of the event from news or media companies' perspectives.

\begin{figure}[t]
  \includegraphics[width=0.46\textwidth]{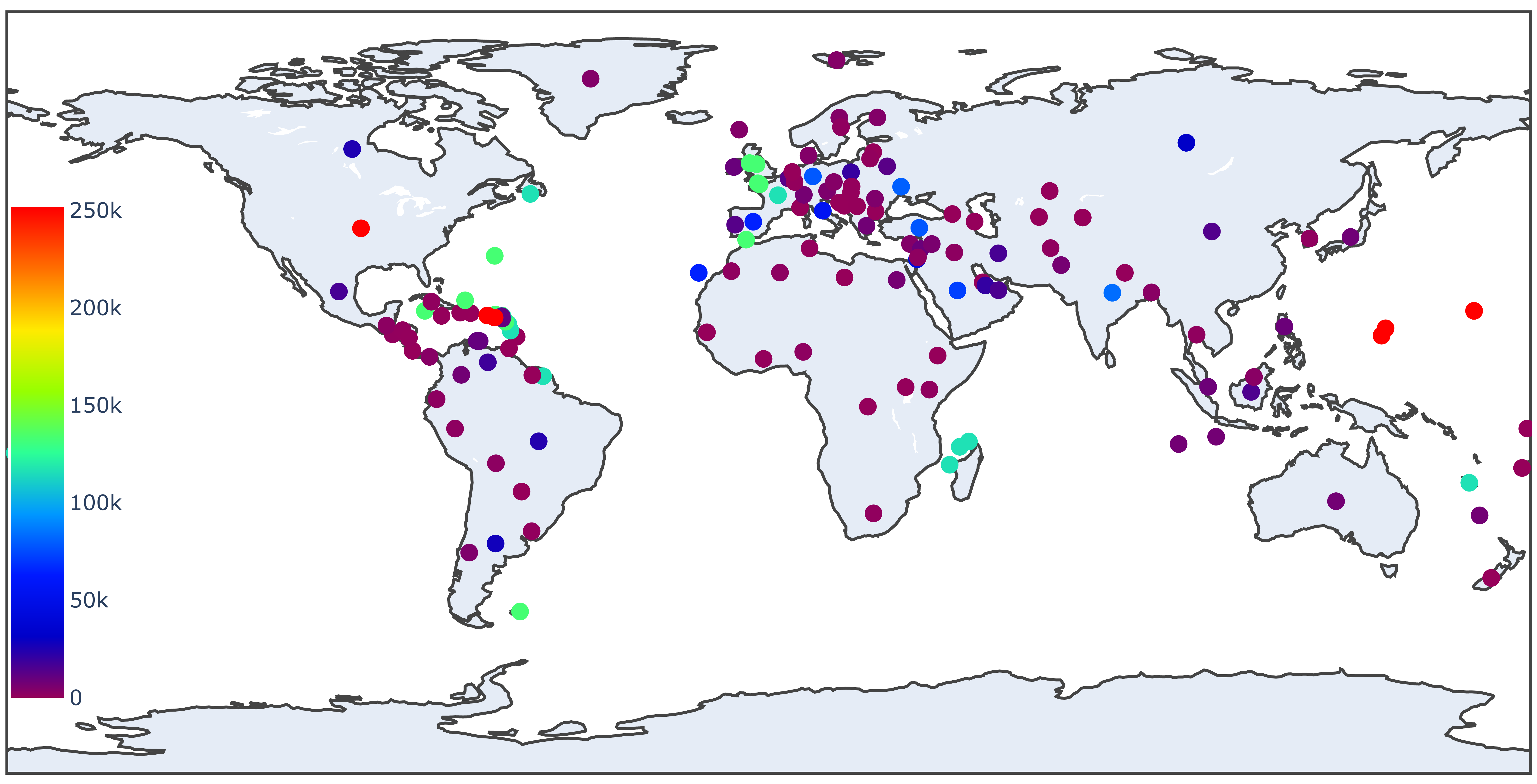}  
  \caption[width=\textwidth]{Distribution of tweets across countries}
  \label{fig:country_map}
  \vspace{-0.6cm}
\end{figure}

In response to this need, we present a multimedia dataset composed of tweets with images from news media channels worldwide that pertain to the Russo-Ukrainian war. This dataset spans a period of February 2022 - May 2023. The dataset is unique in its global scope, encompassing tweets in \num{60} languages and from different parts of the world (see Figure~\ref{fig:country_map}). We downloaded the images for tweets that include them. Additionally, we extracted information about the stance, sentiment, prominent entities \& concepts that occur in tweets, and classified visual concepts in images of tweets to be able to answer questions about the discourse on the ongoing event: who says what (prominent entities), who stands (stance) where on what aspect (prominent concepts), how are the aspects portrayed (sentiment), and finally what is visually portrayed. This comprehensive collection of data allows for a nuanced analysis of the global coverage of the Russo-Ukrainian war, providing insights into how different regions and cultures perceive and report on the conflict. The dataset will serve as a valuable resource for researchers interested in media studies, conflict analysis, and international relations, facilitating a deeper understanding of the global narrative surrounding the Russo-Ukrainian conflict.

\section{Related Work}\label{sec:related_work}

The Russo-Ukrainian conflict has been the subject of extensive research in the realm of social media analysis. Several datasets have been curated to study various aspects of this conflict. \citet{park-etal-2022-challenges} introduced the \textit{VoynaSlov} dataset, a collection of over \num{38} million posts from Russian media outlets on Twitter and VKontakte, to analyze information manipulation. The dataset is used to analyze media effects and to discuss challenges and opportunities in NLP research on information manipulation campaigns. Similarly, \citet{Alyukov2023warmm} studied the manipulation of information shared on Russian social media platforms. \citet{russian_propaganda_social_media} focused on analyzing tweets shared to support of Russia's stance on the conflict and concluded that many bot accounts were deployed to disseminate propaganda on Twitter. Similar study by \citet{Pierri_2023} also focused on analyzing shared propaganda and misinformation on Facebook and Twitter.
\citet{vahdat2023russia} investigated English tweets on the Russia-Ukraine war to analyze trends reflecting users’ opinions and sentiments regarding the conflict. 
\citet{caprolu2022characterizing} built a dataset for the conflict by collecting more than \num{5.5} million tweets related to the subject and performed Aspect-Based Sentiment Analysis (ABSA) to characterize the sentiment about the conflict shared on Twitter in the English-speaking world. Similarly, \citet{shevtsov2022twitter} provided a dataset of \num{57.3} million tweets from \num{7.7} million users and provided a glimpse of volume and sentiment trends in the data. \citet{xu2023sentiment} conducted a study on sentiment analysis using Long Short-Term Memory (LSTM) and Sastrawi on an Indonesian Twitter dataset. \citet{dvzubur2022semantic} focused on tweets related to the Russo-Ukrainian conflict and combined sentiment and network analysis approaches to produce various important insights into the discussion of the conflict. 
\citet{chen2022tweets} have collected more than 600 million tweets between Feb 2022-Feb 2023 while \citet{zhu2022reddit} analyzed communities (subreddits) on Reddit related to the event.

Different from these aforementioned datasets, our focus lies on extracting the global news reporting and discourse on the topic from the perspective of news or media companies around the world. We collected tweets only from user accounts that are tied to a specific media company. Expanding on existing datasets and approaches on the same event, we additionally included both textual and visual concepts that are essential for discourse analysis on multimedia data such as stance, sentiment, and prominent entities as well as concepts in the text and image content.

\section{Dataset}\label{sec:dataset}

\begin{table}[t!]
 \small
  \centering
  \caption{Distribution of collected tweets across \num{60} languages}
  \begin{tabular}{cc}
    \begin{minipage}{0.45\linewidth}
      \centering
      \begin{tabular}{|l|c|}
\hline
\textbf{Language} & \textbf{Count} \\ \hline
English& 581469\\ \hline
Arabic& 169682\\ \hline
Spanish& 163558\\ \hline
French& 111041\\ \hline
German& 74116\\ \hline
Italian& 54336\\ \hline
Turkish& 51274\\ \hline
Dari& 42138\\ \hline
Ukrainan& 41482\\ \hline
Russian& 38198\\ \hline
Portuguese& 38180\\ \hline
Hebrew& 21133\\ \hline
Hindi& 16681\\ \hline
Polish& 16427\\ \hline
Catalan& 16068\\ \hline
Indonesian& 12853\\ \hline
Dutch& 12737\\ \hline
Greek& 11460\\ \hline
Korean& 8274\\ \hline
Japanese& 6143\\ \hline
Urdu& 5263\\ \hline
Romanian& 4944\\ \hline
Denmark& 4103\\ \hline
Norwegian& 3643\\ \hline
Swedish& 2674\\ \hline
Finnish& 2646\\ \hline
Czech& 2226\\ \hline
Bengali& 2116\\ \hline
Gujarati& 1358\\ \hline
Thai& 1262\\ \hline

      \end{tabular}

    \end{minipage}
    &
    \begin{minipage}{0.45\linewidth}
      \centering
      \begin{tabular}{|l|c|}
\hline
\textbf{Language} & \textbf{Count} \\ \hline
Malay& 1007\\ \hline
Telugu& 900\\ \hline
Tamil& 832\\ \hline
Oriya& 761\\ \hline
Tagalog& 613\\ \hline
Basque& 540\\ \hline
Slovakian& 517\\ \hline
Vietnamese& 368\\ \hline
Estonian& 274\\ \hline
Belarussian& 254\\ \hline
Marathi& 213\\ \hline
Nepali& 191\\ \hline
Azeri& 160\\ \hline
Bulgarian& 98\\ \hline
Chinese& 90\\ \hline
Assamese& 78\\ \hline
Pashto& 72\\ \hline
Hungarian& 67\\ \hline
Croatian& 59\\ \hline
Bosnian& 58\\ \hline
Luxembourgish& 44\\ \hline
Lithuanian& 33\\ \hline
Slovenian& 20\\ \hline
Swahili& 17\\ \hline
Welsh& 15\\ \hline
Tajik& 15\\ \hline
Ganda& 14\\ \hline
Kazakh& 13\\ \hline
Latvian& 12\\ \hline
Irish& 12\\ \hline
      \end{tabular}
    \end{minipage}
  \end{tabular}
  
  \label{tab:lang_lang_stats2}
\end{table}

In this section, we provide details about the collection, filtering and processing steps. Our dataset covers \num{60} languages with total of \num{1524826} tweets, out of which \num{306295} have images. These tweets were posted by news or media companies around the world. The source code and the dataset are publicly available~\footnote{
\url{https://github.com/sherzod-hakimov/ru-ua-news-discourse-twitter}
}.

\subsection{Data Collection}\label{subsec:data_collection}

\textbf{Extraction of Twitter Handles}: We have used Wikidata~\cite{wikidata} to query for the news companies, their countries, and Twitter handles~\footnote{Wikidata query for news companies' Twitter handles and countries: \url{https://t.ly/XEp6}}. The query returned \num{14587} news/media companies and their respective countries.

\noindent
\textbf{User Account Verification}: We used Twitter API to check which of the returned user accounts are \textit{verified}, which was done in March 2022. We kept verified accounts with at least \num{100000} and non-verified accounts with at least \num{5000} followers. In total, we ended up with \num{1795} verified and \num{1343} non-verified accounts.

\noindent
\textbf{Querying of Tweets}: Using the extracted and filtered Twitter handles of news companies from the previous steps, we have queried all tweets posted on these accounts between Feb 1st, 2022 - May 31st, 2023. In total, we have collected around \num{47} million tweets.

\subsection{Data Filtering}

\textbf{Keywords}: We manually curated a list of keywords in English that identify with the Russo-Ukrainian conflict (similar to \citet{chen2022tweets}). Later, we translated the keywords into other languages, which are spoken in countries where news companies are located, using \textit{Google Translate} and \textit{DeepL}. The full list is provided\footnote{\url{https://github.com/sherzod-hakimov/ru-ua-news-discourse-twitter/blob/main/resources/language_resources.jsonl}}. 

\noindent
\textbf{Filtering}: To remove irrelevant tweets, we applied two filtering steps. The first step is based on 
removing tweets that do not include the target keywords in the respective languages. The second step is based on prompting Flan-T5-large model~\cite{flan-t5} to check whether the tweet text is related with the topic of interest. It was done by prompting the language model to output \textit{yes/no} by answering the following question ``Does the following tweet mention or relate to Russia-Ukraine conflict?''. 
After applying these steps, we ended up with a total of \num{1524826} tweets for \num{60} languages. 
The distribution of tweets across languages is available in Table~\ref{tab:lang_lang_stats2}.

\begin{figure*}[t]
  \includegraphics[width=\textwidth]{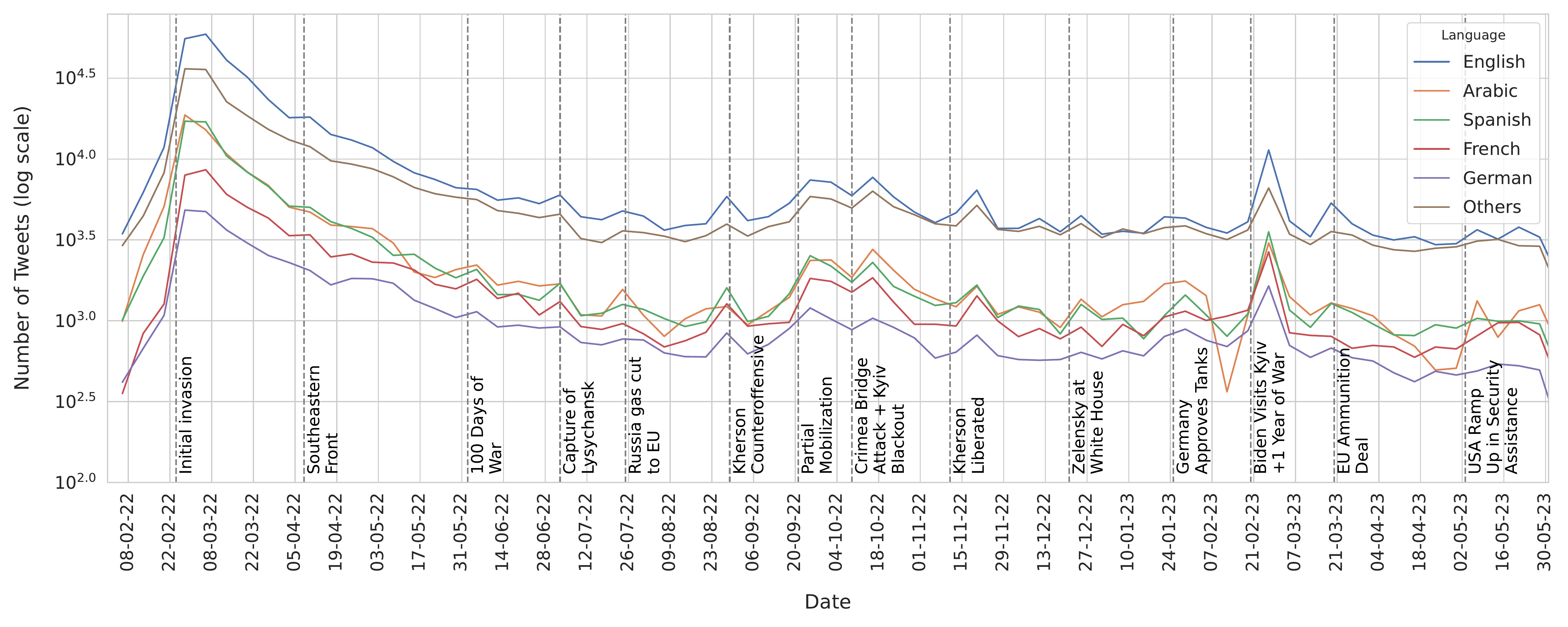}  
  \caption[width=\textwidth]{Distribution of tweets across languages over timeline. Certain prominent events are added manually (dashed lines).}
  \label{fig:lang}
  \vspace{-0.3cm}
\end{figure*}

\begin{figure*}[!htp]
  \includegraphics[width=\textwidth]{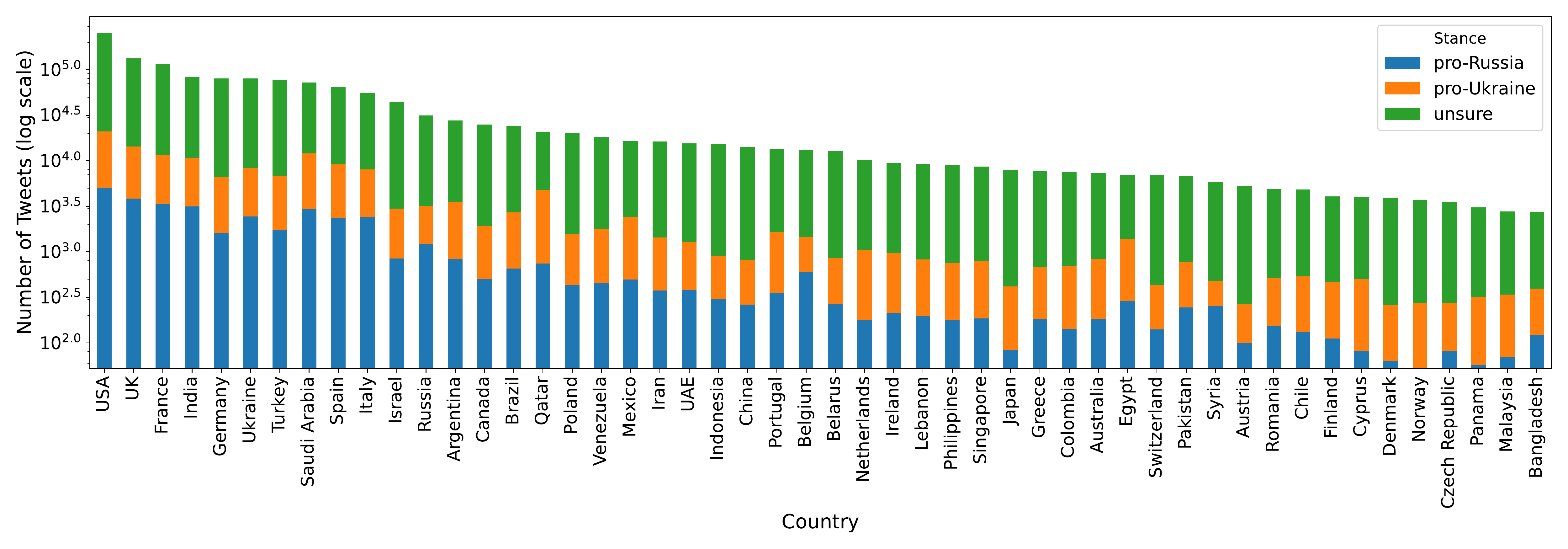}  
  \caption[width=\textwidth]{Distribution of stance in tweets across top 50 countries}
  \label{fig:stance}
\end{figure*}

\begin{figure*}[!htp]
  \includegraphics[width=\textwidth]{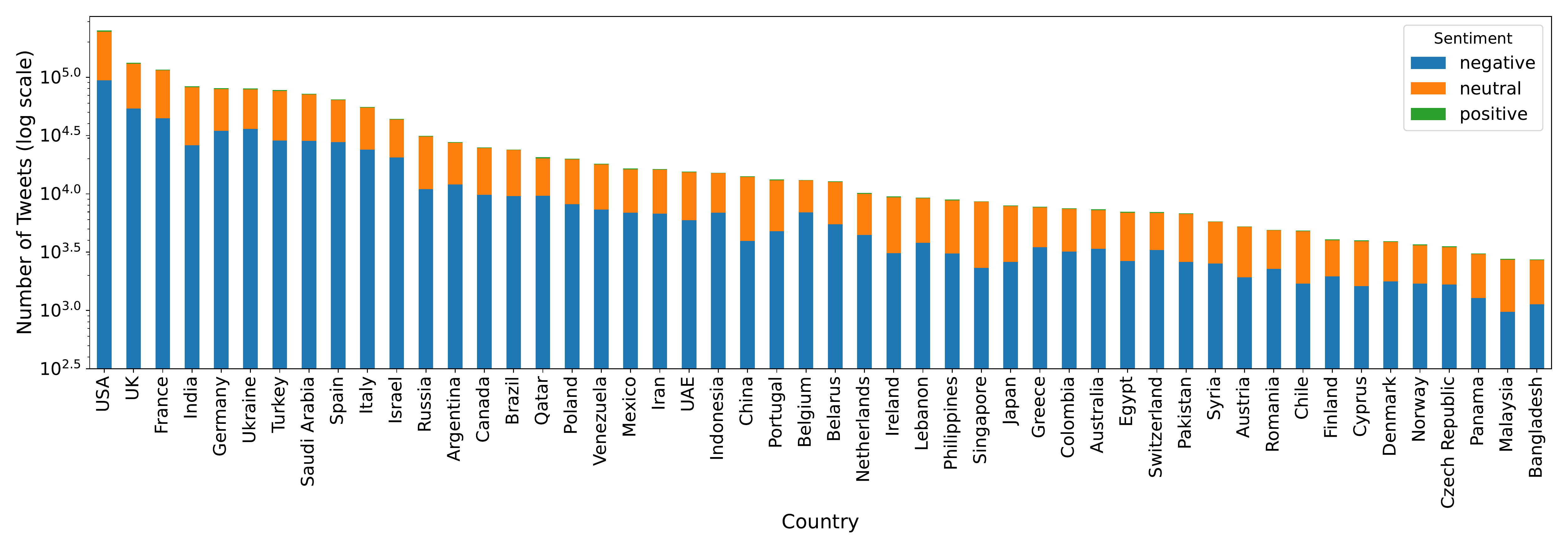}  
  \caption[width=\textwidth]{Distribution of sentiment in tweets across top 50 countries}
  \label{fig:sentiment}
\end{figure*}

\begin{figure*}[!htp]
  \centering
  \begin{subfigure}[b]{0.49\textwidth}
  \includegraphics[width=1.0\linewidth]{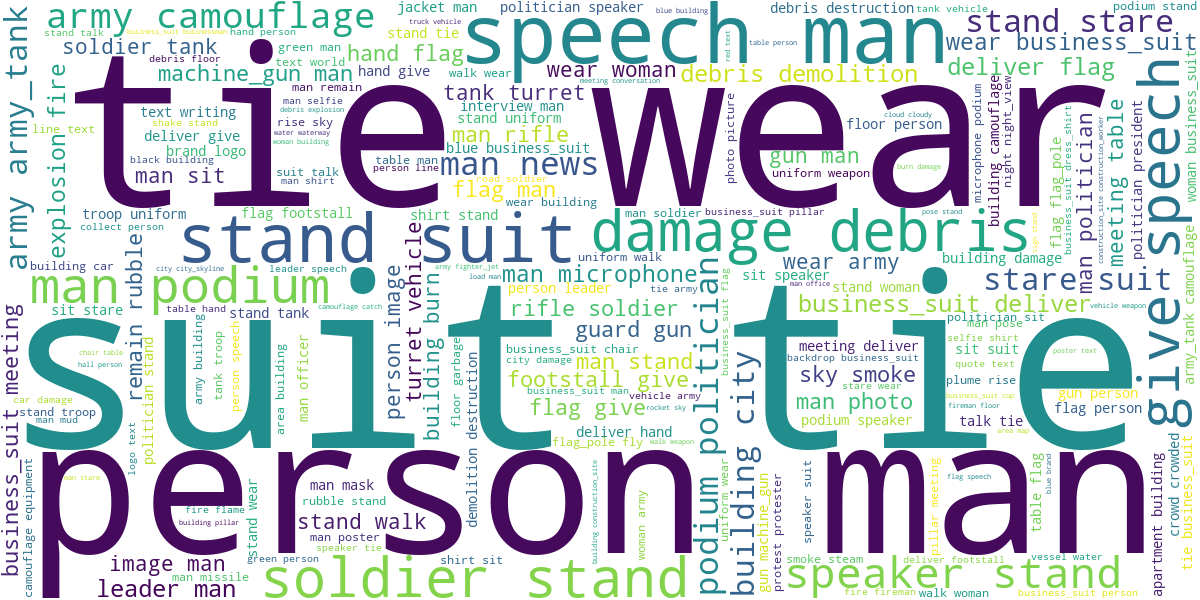}
    \caption{Word cloud of concepts extracted from images}
    \label{fig:image_word_cloud}
  \end{subfigure}
  \begin{subfigure}[b]{0.49\textwidth}
  \includegraphics[width=1.0\linewidth]{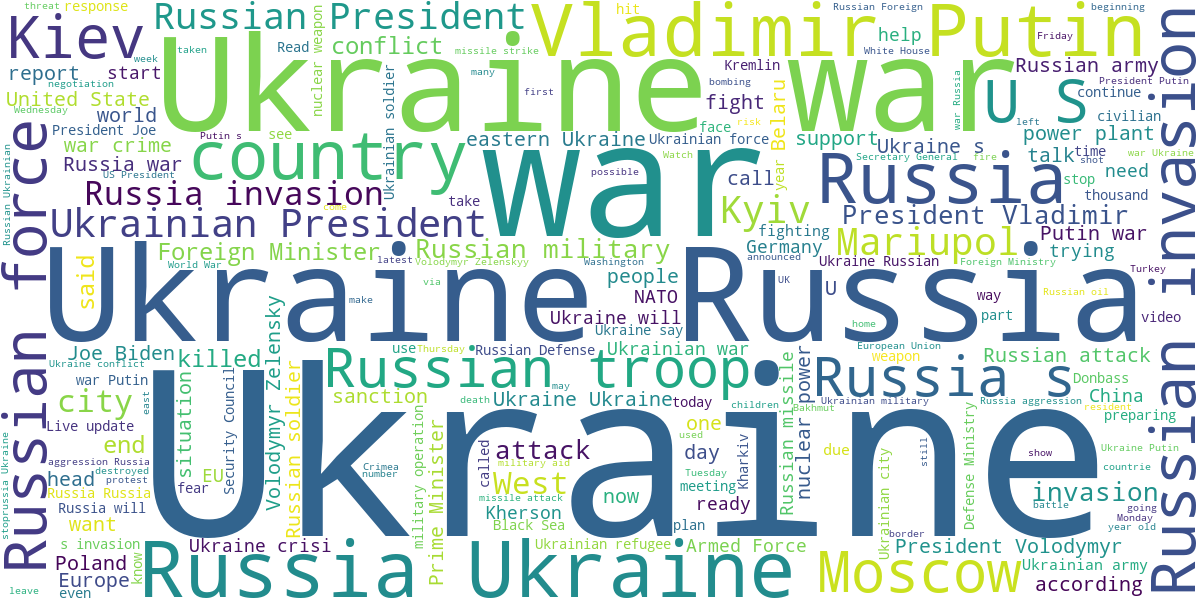}
    \caption{Word cloud of concepts extracted from tweet text (English translations)}
    \label{fig:text_word_cloud}
  \end{subfigure}
  \caption{Word cloud of concepts in text and images}
  \label{fig:word_cloud}
\end{figure*}

\subsection{Data Processing}\label{subsec:data_processing}
The main motivation for creating this multimedia dataset is to be able to analyse the discourse from the news or media companies' perspectives on where do they stand and how they reflect on which aspects. We process both textual and visual content of each tweet. We downloaded all media links in tweets that are of type ``photo'', which resulted in \num{306295} images. Next, we applied pre-trained models to extract class labels of interest from both textual or visual content.

\noindent
\textbf{Language detection}: Each tweet by default is assigned a language tag. In cases where the language tags were \textit{undefined}, we used a language identification model~\footnote{\url{https://huggingface.co/papluca/xlm-roberta-base-language-detection}} to classify them.

\noindent
\textbf{Translation}: To deal with the disparity of NLP tools to apply on \num{60} languages, we decided to translate all non-English tweets into English to be able to use the same pre-trained models without having to deal with the trade-off of having multilingual models vs. performance. We used the pre-trained machine translation model called No Language Left Behind~\cite{nllb2022} (version \textit{nllb-200-1.3B)}.

\noindent
\textbf{Normalization}: We replaced all user and URL mentions with placeholders (<USER\_MENTION>, <URL>). 

\noindent
\textbf{Sentiment}: We used the pre-trained model~\cite{loureiro-etal-2022-timelms} for sentiment analysis that outputs probabilities for \textit{positive}, \textit{negative}, or \textit{neutral}\footnote{\url{https://huggingface.co/cardiffnlp/twitter-roberta-base-sentiment-latest}}.

\noindent
\textbf{Stance}: Stance detection essentially reveals whether a given premise (tweet text) is aligned with a hypothesis. Hypotheses in this study are related to identifying whether the content is against or in favour of concepts such as \textit{military conflict}, \textit{war}, \textit{Russia}, \textit{Ukraine}. We used following hypotheses: ``This statement is in favour of Russia'', ``This statement is against Russia'', ``This statement is against Ukraine'', ``This statement is in favour of Ukraine'', ``This statement is in favour of war'', ``This statement is against war'', ``This statement is in favour of military conflict'', ``This statement is against military conflict''. We used a pre-trained BART model~\cite{DBLP:journals/corr/abs-1910-13461} that is fine-tuned on Multi-Genre Natural Language Inference~\cite{N18-1101} dataset~\footnote{\url{https://huggingface.co/facebook/bart-large-mnli}}.

\noindent
\textbf{Prominent entities, concepts}: We used the Stanza model~\cite{qi2020stanza} to extract Part-of-Speech (POS) tags, named entities from all tweets. The prominent concepts can be constructed from \textit{noun} POS tags whereas entities are directly extracted by the model.

\noindent
\textbf{Visual concepts}: We applied an image recognition model~\cite{huang2023openset} (version \textit{RAM++}) on images to extract detected concepts in images.

\section{Analysis}

\subsection{Preliminary Findings}

\textbf{Languages, events \& number of tweets}: Figure~\ref{fig:lang} depicts the number of tweets posted across top-5 languages (English, Arabic, Spanish, French, German) and others combined over the timeline. We can observe that the event has attracted similar attention across all compared languages and the peaks in a certain time frame also correlate among the compared groups. In addition, the plot shows some of the key events in the war that can potentially explain the increase in tweets at a particular time.

\textbf{Stance}: Figure~\ref{fig:stance} shows the distribution of stance across the top-50 countries. The stances are either \textit{pro-Russia}, \textit{pro-Ukraine}, or \textit{unsure}. The values are calculated by merging the stance outputs for hypotheses mentioned in Section~\ref{subsec:data_processing}. All confidence values equal or higher than \num{0.9} for hypotheses ``This statement is in favour of Russia'', ``This statement is in favour of war'', ``This statement is in favour of military conflict'' are grouped under \textit{pro-Russia} category. Similarly for hypotheses ``This statement is against Russia'', ``This statement is in favour of Ukraine'', ``This statement is against war'', ``This statement is against military conflict'' with confidence values equal or higher than \num{0.9} are grouped under \textit{pro-Ukraine}. The remaining data points are simply merged under \textit{unsure} category. As a result, the tweets are categorized by stances as follows: \num{89}\% unsure, \num{7.3}\% pro-Ukraine, \num{2.7}\% pro-Russia.

\textbf{Sentiment}: The distribution of sentiment across the dataset is as follows: \num{58}\% neutral, \num{40}\% negative, and \num{2}\% positive (see Figure~\ref{fig:sentiment}).

\textbf{Visual concepts}: Figure~\ref{fig:image_word_cloud} shows the most prominent concepts that appear in images. As we can observe, the concepts such as \textit{tie wear, suit, tie, person, man} being the most common on top of concepts such as \textit{damage, debris, man, podium, give speech, army camouflage, army tank}. It suggests that most images depict either people giving speech in a formal clothing or the debris of buildings and military vehicles and related concepts. 

\textbf{Textual concepts}: Figure~\ref{fig:text_word_cloud} shows word cloud generated from tweet text (or translations for languages other than English). As we can see, the most prominent words refer to two countries, their presidents and other related countries.

\subsection{Potential Use-cases}
\textbf{Comparative political science studies and influence analysis:} Researchers can compare and contrast global perceptions of the Russo-Ukrainian conflict with other similar geopolitical events. This allows for a deeper understanding of the nuances in global reactions to different conflicts and can help explain factors that influence these reactions. Similar to the analysis of retweeted country and retweeter country in ~\cite{chen2022tweets}, an analysis of the country as well as media channels can be performed to identify key sources of news and influence, and how they shape the global conversation on the conflict. Such analysis can reveal underlying patterns of power dynamics and geopolitical influence in shaping the narrative around international conflicts.

\noindent\textbf{Linguistic studies, journalism and media studies:} The tweets in 60 different languages can help us see how media channels talk about the conflict across cultures and languages. It provide insights into the region and culture-specific idioms, metaphors, and linguistic structures used in reporting and discussing the conflict. The dataset also offers a means to examine the framing of the conflict by different media outlets worldwide. It can facilitate the identification of potential biases, political stances (via stance detection), and varying reportage styles, contributing to a more comprehensive understanding of the global media landscape and its impact on conflict narratives.

\noindent\textbf{Social media dynamics, fake news and misinformation studies:} The dataset allows for a thorough investigation of how narratives spread and evolve over time. It offers a real-world case study on information propagation, topic lifespan, and the dynamics of virality on social media platforms. It is also a key resource for studying misinformation, often prevalent in conflicts, helping identify misinformation patterns. The inclusion of verified and non-verified accounts adds an extra dimension to fake news analysis.

\section{Conclusion}\label{sec:conclusion}

We present a new multimedia dataset that focuses on unveiling the global narrative of the ongoing Russo-Ukrainian conflict by analysing the tweets posted by news or media companies around the world. We collected the tweets between February 2022 - May 2023 and applied certain filtering and NLP pipelines to acquire around 1.5 million tweets with their images in \num{60} different languages. Our dataset includes processed tags for each tweet to be able to answer questions such as who says what (prominent entities), who stands (stance) where on what aspect (prominent/visual concepts), and how are the aspects portrayed (sentiment). The existence of such a dataset will serve as a valuable resource for researchers aiming to study the global narrative from various aspects.\\
\textbf{Limitations}: The main limitation of the approach is reliance on tools that are pre-trained on English data since all tweet text were translated to English. The main reason behind this is not having such language tools for all languages of interest.\\
\textbf{Ethical Considerations}: The publicly shared dataset will be composed of only tweet IDs and the processed information described in Section~\ref{subsec:data_processing}. The full tweet text and images can be only be released upon formal request only for scientific purposes.

\bibliographystyle{acmbib}
\bibliography{references}

\end{document}